\pdfoutput=1

\documentclass[11pt]{article}

\usepackage[preprint]{acl}

\usepackage{times}
\usepackage{latexsym}
\usepackage{amsmath,amssymb}
\usepackage{graphicx}
\usepackage{bm}
\usepackage{subcaption}
\usepackage{cleveref}
\crefformat{section}{\S#2#1#3}
\crefformat{subsection}{\S#2#1#3}
\crefformat{subsubsection}{\S#2#1#3}
\crefformat{appendix}{\S#2#1#3}
\usepackage{multirow} 
\usepackage{caption}
\usepackage{stfloats}
\usepackage[T1]{fontenc}
\usepackage[utf8]{inputenc}
\usepackage{microtype}
\usepackage{inconsolata}

\title{Stories that (are) Move(d by) Markets: A Causal Exploration of Market Shocks and Semantic Shifts across Different Partisan Groups}

\author{Felix Drinkall*, Stefan Zohren*, Michael McMahon\dag, Janet B. Pierrehumbert*\ddag \\
    *Department of Engineering Science, University of Oxford \\
    \dag Department of Economics, University of Oxford \\
    \ddag Faculty of Linguistics, University of Oxford \\
    \texttt{felix.drinkall@eng.ox.ac.uk}}

\begin{document}
\maketitle
\begin{abstract}
Macroeconomic fluctuations and the narratives that shape them form a mutually reinforcing cycle: public discourse can spur behavioural changes leading to economic shifts, which then result in changes in the stories that propagate. We show that shifts in semantic embedding space can be causally linked to financial market shocks -- deviations from the expected market behaviour. 
Furthermore, we show how partisanship can influence the predictive power of text for market fluctuations and shape reactions to those same shocks. We also provide some evidence that text-based signals are particularly salient during unexpected events such as COVID-19, highlighting the value of language data as an exogenous variable in economic forecasting. Our findings underscore the bidirectional relationship between news outlets and market shocks, offering a novel empirical approach to studying their effect on each other.

\end{abstract}

\section{Introduction}

Within economics and finance, a complex dialectic emerges between market events and the narratives that drive them. Narratives can drive changes in beliefs which drive behavioural changes that lead to macroeconomic shifts; but equally, narratives can be shaped by the observable behavioural changes and the economic data. Whilst the link between narratives and economic actions has been the subject of many influential works \cite{akerlof2010animal, shiller2017narrative}, the two-way endogeneity makes this complex relationship very hard to model. One popular approach is to focus on how specific historical narratives persisted and their effect on the world \cite{margalit2019political,Tangherlini2020,Ash_Gauthier_Widmer_2024}. However, developing a predictive system capable of understanding the causal influence between narratives and the economy, and vice versa, is a far more challenging task -- one that this paper aims to address.

Economics has long incorporated exogenous data \cite{einav2014economics} -- information which exists outside of the economic system itself. 
Linguistic behaviour from news and social media can act as a proxy for consumer beliefs which adds nuance to predictive feature sets, contingent on the text encoding methodology. Existing efforts to model the relationship between the real world and written or spoken text have tended to focus on more interpretable and simplistic text representations such as lexicons \cite{loughran2011liability} and sentiment \cite{math10132156}. Some efforts have centered around extracting narrative-form structures in the form of subject-verb-object triplets to understand the nature and directionality of what is contained in the text \cite{piper2021narrative, Ash_Gauthier_Widmer_2024}. However, narrative extraction can suffer from sparsity issues limiting its use in predictive applications. The recent progress in transformer-based LLMs potentially solves this problem by looking at the text in a distributional rather than an extractive manner - perhaps another example of the "Bitter Lesson" \cite{sutton2019bitter}. LLMs are able to encode more granular text features than extractive methods. This paper leverages the representational power of transformer-based LLMs to understand the bidirectional relationship between text and the macroeconomy.

Different narratives can circulate about the same economic event, with these differences sometimes stemming from partisan differences \cite{Epstude2022, HOBOLT_LAWALL_TILLEY_2024}. News outlets of various political orientations — centrist, right-wing, and left-wing — can provide contrasting interpretations of identical events \cite{doi:10.1073/pnas.2013464118}. To better understand the relationship between news outlets and the economy, it is important to examine how different partisan groups influence and react to financial markets. Existing research typically offers broad explanations for how events — such as financial crises or elections — shape subsequent political reactions \cite{Klüver_Spoon_2016, margalit2019political}. This paper looks at political groups separately and on a daily basis to understand the differences between how partisan groups respond to and precede market shocks.

Feedback loops are of great interest to researchers and practitioners attempting to predict virality. News interest can increase attention which can itself increase the news interest, entering a self-reinforcing cycle. Viral feedback loops can exist in multiple domains: tourism \cite{MURRAY1997513}, social media \cite{MetzlerGarcia2024}, consumer products \cite{NGUYEN201939}, and other areas of the financial markets \cite{semenova2024wisdom}. This paper attempts to quantify feedback loops from a linguistic diachronic change perspective paving a path for other domains.

This paper makes the following contributions:

\begin{itemize}
    \item First paper to show that shifts in semantic embedding space can be causally linked to measurable real-world shocks.
    \item We conduct an analysis of how partisanship affects the ability of text to predict market shocks and the reaction to market shocks.
    \item We show that feedback loops exist between the market and news outlets.
    \item We find evidence that text is an important exogenous variable in violent economic events such as COVID-19.
\end{itemize}
\vspace{-0.1cm}

\section{Literature Review}
\vspace{-0.1cm}

\subsection{Text-based Forecasting}

Text-based forecasting explores the predictive power of textual data such as news articles \cite{peng-jiang-2016-leverage, 10.1145/3533018, wang2024newsforecast, rahimikia2024r}, social media \cite{10.1145/2542182.2542190, iso-etal-2016-forecasting}, or corporate reports \cite{kogan-etal-2009-predicting}. By extracting linguistic features such as keywords \cite{iso-etal-2016-forecasting}, sentiment \cite{math10132156}, raw embedding representations \cite{sawhney-etal-2020-deep} or high-density embedding clusters \cite{drinkall-etal-2025-forecasting, drinkall2025financialregression}, researchers have shown that it is possible to enhance traditional forecasting models using text-derived information. The findings of these papers imply there is some causal interaction between textual data and the economy. Yet to our knowledge, no papers have explored the bidirectional nature of this relationship; understanding the effect that markets have on news discourse. 
\vspace{-0.1cm}

\subsection{Diachronic Shifts}
\label{sec:diachronic}

Diachronic analysis \cite{10.1145/3672393} examines how linguistic phenomena change over time. The process begins by aggregating time-stamped information to create time-specific representations. The aggregation process can be done using averaging \cite{10.1007/978-3-030-72610-2_13} or clustering \cite{10.1145/3366424.3382186}, and is considered a design choice dependent on what one is measuring; with averaging suiting cases which aim to detect the dominant shifts. Diachronic shifts tend to be measured using a distance metric between the point-in-time representations. There are several distance metrics that are used in the literature: Cosine Distance (CD) \cite{martinc-etal-2020-leveraging, horn-2021-exploring}, Average Pairwise Distance (APD) \cite{giulianelli-etal-2020-analysing, kudisov-arefyev-2022-black}, Hausdorf Distance (HD) \cite{Wang2020} etc. The standard distance metric is CD -- which we use in this paper -- since APD is more commonly used for polysemy detection \cite{keidar-etal-2022-slangvolution} and HD is sensitive to outliers \cite{Wang2020}. The reciprocal of CD, Cosine Similarity, is widely used in other NLP applications as a robust and efficient similarity metric \cite{thongtan-phienthrakul-2019-sentiment, reimers-gurevych-2019-sentence, yamagiwa-etal-2025-revisiting}. 
\vspace{-0.1cm}

\subsection{Causality}
\label{sec:causality}

\subsubsection{Causality in Time-series}
\label{sec:caus_in_TS}

Causality in time-series can be inferred using methods like transfer entropy \cite{barnett2009granger}, convergent cross mapping \cite{ye2015distinguishing} and dynamic Bayesian networks \cite{ghahramani1997learning}; these techniques model non-linear effects but are poor at isolating individual lag contributions. Granger causality \cite{702ab909-8cb1-3c30-a5f1-ab4517d6cf1c} tests whether past values of one variable can help to predict the future value of another, enabling analysis of individual lag contributions. It does not imply “causation” in a philosophical sense but identifies temporal precedence and predictability. If the inclusion of past values of variable $X$ significantly improves the prediction of variable $Y$, then $X$ is said to "Granger cause" $Y$. Non-linear regression models can be used in a Granger causality test to model non-linear relationships \cite{marinazzo2008kernel}.
\vspace{-0.1cm}

\subsubsection{Causality in Text}
\label{sec:cause_in_text}

There are three main research areas that explore "causal" interactions in NLP. Firstly, autoregressive generation which underpins modern generative LLMs is sometimes referred to as "causal language modelling" \cite{Radford2018, Radford2019, Brown2020}, but this area is not related to the identification of causal relationships with exogenous data. Secondly, causal relation extraction \cite{ning-etal-2018-joint, dasgupta-etal-2018-automatic-extraction, Yang2021ASO} identifies causal relationships between clauses. The final causal NLP research area which is of interest to this paper creates a link between text and observable phenomena in the real world. Text-derived features can be used in causal inference where text is the treatment variable - what causes the outcome - for a temporally static outcome. Past work has found success in medical records \cite{Zeng2021UncoveringIP}, politics \cite{fong-grimmer-2016-discovery}, and mental health \cite{zhang2020quantifying}. This research direction is mature and has developed solutions that encode text with topic models \cite{roberts2020adjusting} as well as text embeddings \cite{Veitch2019AdaptingTE}. However, there has been little work exploring the predictive causal relationship between text and time-series information. Existing methods do not link changes in an LLM-derived semantic space with observable time-series, instead focusing on keywords \cite{balashankar-etal-2019-identifying, Maisonnave2022CausalGE}, lexicons, and LDA topics \cite{kang-etal-2017-detecting}, making our framework novel. Further to this, no techniques measure the temporal causality of observable time-series on text.
\vspace{-0.1cm}

\section{Datasets}
\label{sec:datasets}
\vspace{-0.1cm}

\subsection{News Dataset}
\label{sec:news_dataset}

We use the news dataset from \citet{spliethoever2022}, which spans a large time period from 2000 to 2022 and includes the political orientation labels from \href{https://www.allsides.com}{AllSides.com}. The news outlets included in this paper, alongside their partisan label, are listed in \cref{app:news_outlets}. We use data from 01/07 to 02/22 to ensure sufficient coverage.

\vspace{-0.1cm}

\paragraph{Text Representation.}Let $S$ be the set of synonyms for the terms \{\emph{prices}, \emph{inflation}, \emph{labor}, \emph{growth}\}. Let $\mathcal{O}$ be the set of news outlets, and let $t$ index each day in the dataset. Synonyms were isolated by identifying the top 50 most similar words in the vocabulary of \textit{all-mpnet-base-v2}, and then removing problematic polysemic words.
\vspace{-0.1cm}

\paragraph{Article Selection and Embedding.} 
For each day $t$ and outlet $o \in \mathcal{O}$, all articles were stored that referenced at least one synonym in $S$. The embedding $\mathbf{e}_{t,o,k}$ of the $k$-th article from outlet $o$ on day $t$, was obtained using the \textit{all-mpnet-base-v2} model \cite{reimers-gurevych-2019-sentence}. If an article exceeded the model's maximum sequence length, it was split into chunks, and the chunk embeddings were averaged to form a single article representation $\mathbf{e}_{t,o,k}$. An encoder-based model is a natural choice for this task as opposed to a large generative model because bidirectional attention models often performs better on embedding tasks \cite{10.5555/3495724.3497138, muennighoff-etal-2023-mteb, lee2025nvembedimprovedtechniquestraining}.
\vspace{-0.1cm}

\paragraph{Daily Outlet Embeddings.}
Let $N_{t,o}$ be the number of relevant articles published by outlet $o$ on day $t$. The \emph{daily outlet embedding} $\mathbf{E}_{t,o}$ for outlet $o$ on day $t$ was calculated by averaging over the articles:
\[
\mathbf{E}_{t,o} = \frac{1}{N_{t,o}} \sum_{k=1}^{N_{t,o}} \mathbf{e}_{t,o,k}
\]
provided $N_{t,o} > 0$. If $N_{t,o} = 0$, then the embeddings were forward filled:
\[
\mathbf{E}_{t,o} = \mathbf{E}_{t-1,o}
\]
\vspace{-0.1cm}

\paragraph{News Feature.}
The \emph{news feature} for outlet $o$ on day $t$ is the cosine distance between the daily embeddings of consecutive days:
\[
\text{NewsFeature}_{t,o} = 1 \;-\; \frac{\mathbf{E}_{t,o} \cdot \mathbf{E}_{t-1,o}}{\|\mathbf{E}_{t,o}\|\;\|\mathbf{E}_{t-1,o}\|}
\]
This measure quantifies the day-to-day semantic shift for each outlet. We used this inline with the diachronic literature outlined in \cref{sec:diachronic} and because it can be interpreted as a textual shock, mirroring the market shocks that we introduce in \cref{sec:market_shocks}. We only consider outlets where there is an article published on 25\% of the days. The similarity in data format allows us to create a symmetric comparison between news and market data.
\vspace{-0.1cm}

\subsection{Market Shocks}
\label{sec:market_shocks}

To study the effect of evolving market beliefs on narrative, and vice versa, we ideally need high-frequency data. Unfortunately, most economic indices are reported monthly or quarterly, which would miss the transmission of short-term shocks to the economy \cite{chen2015granularity}. News cycles can last a matter of days \cite{10.1145/2531602.2531623}, motivating the need for more granular time series. We therefore focus on market data which has the advantages that (i) it is updated at high frequency on trading days (we use daily data), and (ii) it is understood to reflect the beliefs and information of market participants and their desire to make money means that new information should be reflected quickly in asset prices. Specifically, we use daily asset-price-derived indicators from \citet{cieslak2021common} that capture four distinct types of economic news: \emph{growth}, \emph{monetary}, \emph{common premium}, and \emph{hedging premium} shocks. These shocks are derived by first purging stock prices (S\&P 500) and bond yields of different maturities (two-, five- and ten-year) of their movements that are predictable from systematic relationships between the asset prices. By exploiting economic reasoning, the remaining variation in asset prices can be decomposed into the four distinct types of news:\footnote{Formally, \citet{cieslak2021common} identify the shocks via sign restrictions on a structural vector autoregression (SVAR). The procedure is standard in economics and explained in \cref{app:SVAR}.}

\begin{itemize}
    \item \textbf{Growth \((\omega_g)\)}: Positive growth news raises stock prices and bond yields at short-to-medium maturities.
    \item \textbf{Monetary (tightening) \((\omega_m)\)}: Positive monetary news lowers stock prices and increases short-to-medium bond yields, with a weaker effect at longer maturities.
    \item \textbf{Common premium \((\omega_{cp})\)}: An increase in the discount rate risk premium (often linked to inflation) boosts bond prices, reducing yields, and lowers stock prices, creating negative co-movement between yield changes and stock returns. This effect is stronger at the longer maturities.
    \item \textbf{Hedging premium \((\omega_{hp})\)}: Elevated future cash flow risk depresses stock prices and raises long-term bond yields (safe haven effect), generating positive co-movement between yield changes and stock returns. This effect is stronger at the longer maturities.
\end{itemize}

Growth and monetary policy shocks mainly affect shorter maturities, reflecting mean reversion around recent economic news, whereas risk-premium shocks manifest at longer maturities, driven by longer-term uncertainty. The resultant time-series encapsulate structural deviations from the expected market behaviour which should reflect new beliefs (howsoever arrived at) as earlier information is already reflected in the market prices. The result is four daily orthogonal time-series, each representing the market's evolving interpretation of economic events via their effects on both stock and bond markets.
\vspace{-0.1cm}

\section{Methodology}
\label{sec:methodology}

The goal of the paper is to determine how, if at all, market time-series affect the narrative in the media and narratives in the media drive market prices. That is, whether there is any information in the textual media time-series that is predictive of market time-series, or vice versa. To do this we devised an evaluation tool using Granger causality tests, the steps of which are outlined in the following section:

\begin{enumerate}
  \item \textbf{Lagged Feature Creation}: For each target variable \(y_t\), we construct lagged predictors \(y_{t-\ell}\) and \(x_{t-L}\), \(\ell \in \{1, \dots, L\}\).
  \item \textbf{Estimation and Prediction}: We fit both the base and enhanced models on the available training data and obtained out-of-sample predictions \(\hat{y}_{B,t}\) and \(\hat{y}_{E,t}\), for every news outlet. The enhanced regression model predicts \(\hat{y}_{E,t}\) using \(y_{t-\ell}\) and \(x_{t-L}\), whereas the base model predicts \(\hat{y}_{B,t}\) using only \(y_{t-\ell}\).
  \item \textbf{Error Evaluation}: We compute errors \(e_{B,t}\) and \(e_{E,t}\).
  \item \textbf{Statistical Testing}: We calculate \(d_t = e_{B,t} - e_{E,t}\) and apply paired tests to determine whether differences in prediction errors are significant.
\end{enumerate}

This methodology allows us to isolate and quantify the effect of additional explanatory features relative to a baseline that depends solely on past values of the target variable.
\vspace{-0.1cm}

\subsection{Model Specification}
\label{sec:modelling}

We assess the predictive value of text-based features by comparing two regression model configurations: a \emph{base model} and an \emph{enhanced model}. The base model incorporates only lagged values of the target variable, while the enhanced model adds lagged text-derived or macroeconomic predictors, depending on the domain of the target variable. 
\vspace{-0.1cm}

\paragraph{Base Model.}
Let \(y_t\) be the target variable at time \(t\), and suppose the base model uses \(L\) lagged observations of \(y\) as predictors. We denote the parameter set of the base model by \(\boldsymbol{\theta}_B\):
\begin{equation}
  y_t = f_B\bigl(y_{t-1}, y_{t-2}, \dots, y_{t-L} \mid \boldsymbol{\theta}_B\bigr) + \varepsilon_{B,t}
\end{equation}
\vspace{-0.1cm}

\paragraph{Enhanced Model.}
In the enhanced model, we introduce a single additional predictor \(x_{t-L}\) (e.g., news outlet cosine distances or market shocks) at the same lag level. We denote the parameter set of the enhanced model by \(\boldsymbol{\theta}_E\):
\begin{equation}
  y_t = f_E\bigl(y_{t-1}, \dots, y_{t-L}, x_{t-L} \mid \boldsymbol{\theta}_E\bigr) + \varepsilon_{E,t}
\end{equation}
\vspace{-0.1cm}

\subsection{Regression Models}
\label{sec:reg_models}

We employ two regression models: \emph{Linear Regression} (Linear) and \emph{Kernel Ridge Regression} (KRR) with a radial basis function (RBF) kernel. Comparing the two allows us to determine whether the relationship between the input features and the target variable is linear or non-linear in nature.
\vspace{-0.1cm}

\paragraph{Linear Regression.} In the linear case, we use an OLS loss function that identifies a parameter vector \(\boldsymbol{\beta}\) that minimizes the squared error:

\begin{equation}
    \min_{\boldsymbol{\beta}} \;\| \mathbf{y} - \mathbf{X}\boldsymbol{\beta}\|^2
\end{equation}

where \(\mathbf{X} \in \mathbb{R}^{n\times p}\) denotes the matrix of input features, and \(\mathbf{y} \in \mathbb{R}^{n}\) represents the target values.
\vspace{-0.1cm}

\paragraph{Kernel Ridge Regression.} In the non-linear case - as in \citet{marinazzo2008kernel} - we use an RBF kernel \(K(\mathbf{x}_i, \mathbf{x}_j)\) that projects inputs into a high-dimensional space. The KRR solution \(\boldsymbol{\alpha}\) is found by solving:

\begin{equation}
    \min_{\boldsymbol{\alpha}} \;\bigl\|\mathbf{y} - \mathbf{K}\boldsymbol{\alpha}\bigr\|^2 + \lambda \,\boldsymbol{\alpha}^\top \mathbf{K}\boldsymbol{\alpha}
\end{equation}

where \(\lambda > 0\) controls the trade-off between data fidelity and model complexity.
\vspace{-0.1cm}

\subsection{Error Metrics}

Let \(\hat{y}_{B,t}\) and \(\hat{y}_{E,t}\) be the predictions from the base and enhanced models, respectively, for time \(t\). We denote the prediction errors as:
\[
e_{B,t} \;=\; \hat{y}_{B,t} - y_t
\quad\text{and}\quad
e_{E,t} \;=\; \hat{y}_{E,t} - y_t
\]
We use MSE error to determine the performance of the model configurations:
\begin{align}
  \text{MSE} &= \frac{1}{T}\sum_{t=1}^{T} \bigl(e_{M,t}\bigr)^2
\end{align}
where \(M \in \{B, E\}\) (base or enhanced), and \(T\) is the total number of test observations.
\vspace{-0.1cm}

\subsection{Statistical Significance Test}
\label{sec:stat_test}

We evaluate if $e_{E,t}$ is significantly lower than $e_{E,t}$ with the following null hypothesis:
\[
d_t = e_{B,t} - e_{E,t} \quad \quad H_{0}: \mathbb{E}[d_t] \leq 0
\]
against the one-sided alternative \(H_{1}:\; \mathbb{E}[d_t] > 0\). We conduct the following paired t-test on the sample \(\{d_t\}_{t=1}^T\):
\[
T_{\text{stat}} \;=\; \frac{\overline{d}}{\sqrt{\widehat{\mathrm{Var}}(d)/T}}
\]
where \(\overline{d} = \frac{1}{T}\sum_{t=1}^T d_t\) and \(\widehat{\mathrm{Var}}(d)\) is the sample variance of \(d_t\). A statistically significant positive mean difference \(\overline{d} > 0\) provides evidence that including the additional features reduces prediction errors on average.

\begin{figure*}[b]
    \centering
    \begin{subfigure}[b]{0.49\linewidth}
        \includegraphics[width=\linewidth]{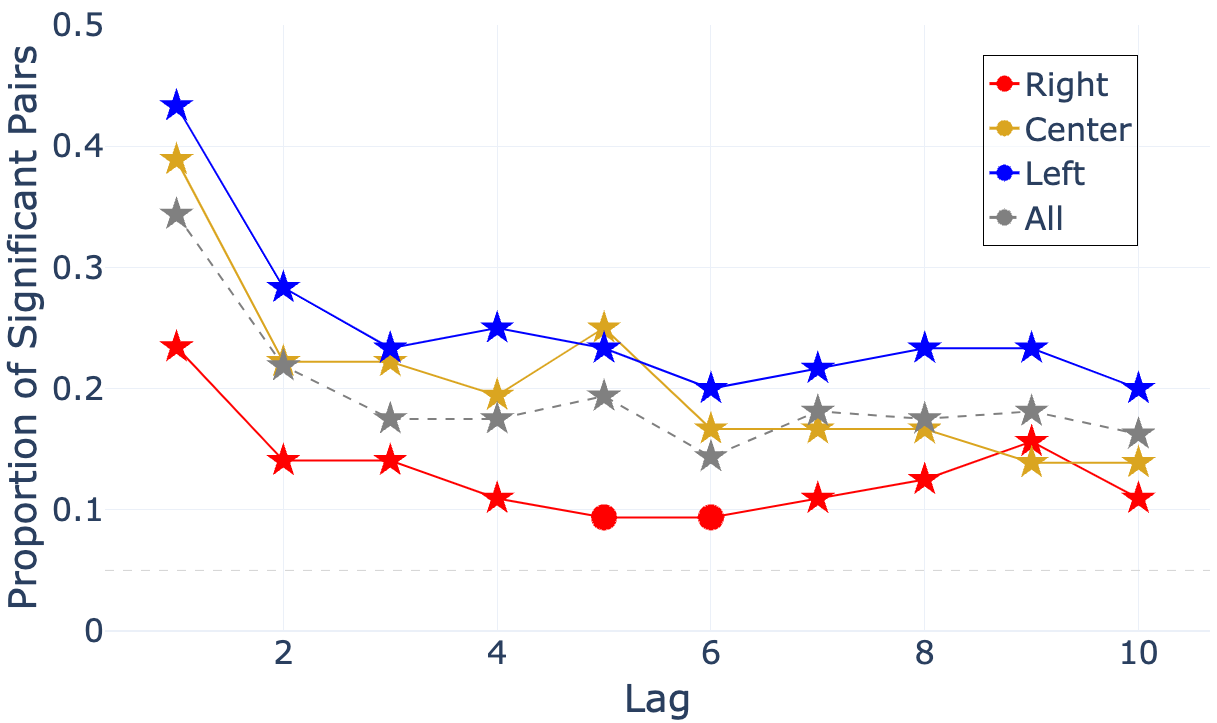}
        \caption{Sign. proportion at different lags for \{$y=\text{text}$, $x=\text{econ}$\}.}
        \label{fig:econ_text_ori}
    \end{subfigure}
    \hfill
    \begin{subfigure}[b]{0.49\linewidth}
        \includegraphics[width=\linewidth]{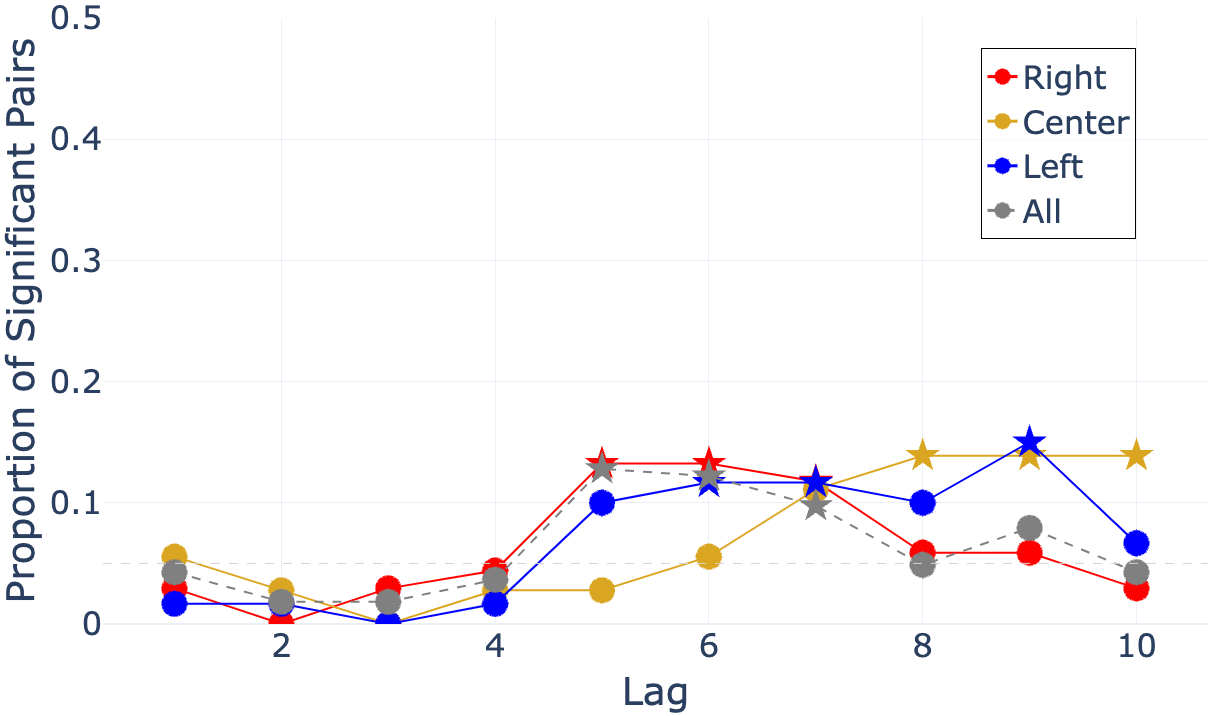}
        \caption{Sign. proportion at different lags for \{$y=\text{econ}$, $x=\text{text}$\}.}
        \label{fig:text_econ_ori}
    \end{subfigure}
    \caption{The proportion of significant pairs for when \{$y=\text{text}$, $x=\text{econ}$\} and \{$y=\text{econ}$, $x=\text{text}$\}. Significant proportions are marked with a star.}
    \label{fig:text_caus_by_ori}
\end{figure*}

\vspace{-0.1cm}

\subsection{P-hacking}
\label{sec:p_hack}

P-hacking arises when a large number of hypothesis tests are conducted, increasing the risk of false positives. Let $\{p_i\}_{i=1}^N$denote the collection of \(p\)-values from \(N\) individual tests, each compared against a standard threshold \(\alpha = 0.05\). Define 
\[
\hat{\rho} \;=\; \frac{1}{N} \sum_{i=1}^N \mathbf{1}\{p_i < \alpha\}
\]
as the empirical proportion of observations deemed significant, to see if this proportion truly exceeds the 5\% expected by chance, we perform a one-sided binomial test under the null hypothesis that \(\hat{\rho}=0.05\). We conclude that a group-level effect is present only if the binomial test p-value is below $\alpha$, indicating that \(\hat{\rho}\) is significantly above $0.05$. 

Moreover, to determine whether a specific pair is significant in \cref{sec:conf} and \cref{sec:circular_causality}, we use a Bonferroni-adjusted threshold \(p_i < \frac{\alpha}{M}\), where \(M\) is the total number of tests \cite{bonferroni1936teoria} across the partisan groups. This procedure reduces the likelihood of false discoveries. In fact, there are multiple studies that suggest that the Bonferroni adjustment is too conservative and it promotes false negative results \cite{perneger1998s, anderson2000null, nakagawa2004farewell}. Given this strict definition of significance, if a pair does emerge as significant we can be more confident that the effect is real. 
\vspace{-0.1cm}

\section{Results}
\label{sec:results}

Unless specified the following results are calculated using a static temporal split, where the training set constitutes 70\% of the samples, the test set made up the final 30\%, and the validation was created using the final 10\% of the training set. The first and most elementary result is outlined in Tab. \ref{tab:lin_vs_nonlin}, which suggests that while all of the configurations exceed the significance threshold, the non-linear regression model identifies more significant pairs than the linear regression model. As a result, we use the non-linear KRR regression model for the remainder of the results in this section.

\begin{table}[h]
    \centering
    \begin{tabular}{c|cc}
         & KRR & Linear \\
         \hline
         $y=\text{text}$, $x=\text{econ}$ & 19.5* & 6.69* \\
         $y=\text{econ}$, $x=\text{text}$ & 6.34* & 6.46* \\
    \end{tabular}
    \caption{Proportion (\%) of significant pairs across all political groups, news outlets and market shocks. An * indicates that the effect is significant. Full data - \cref{app:linvsnonlin}.}
    \label{tab:lin_vs_nonlin}
\end{table}

The first contribution on decomposing the interrelations between narrative and market data, is our finding that market shocks have a greater influence on changes in text than the other way around. This is hardly surprising, but it does imply that news outlets are more reactive than predictive and that the markets are generally efficient (with some exceptions explored below) - pricing in changes before news outlets change their narrative. However, this table averages over the more granular group effects.
\vspace{-0.4cm}

\subsection{Partisan Differences}
\label{sec:res_partisan_diff}

Fig. \ref{fig:text_caus_by_ori} outlines a more nuanced story. Fig. \ref{fig:econ_text_ori} shows that the effect that market shocks have on news discourse approximates a negative exponential. It is also clear that the left-wing and centrist outlets are more reactive to market shocks than right-wing outlets. This relative difference is persistent across time but is more pronounced at shorter time horizons. Fig. \ref{fig:text_econ_ori} shows that at longer time horizons (5-10 days), news outlets do contain some useful predictive information about the market. This suggests that when a significant shift occurs in how a news outlet writes about the economy, there is a delay of around a week before a market correction. We also see that the different political groups display different distributions of significant pairs across the lags - the same as Fig. \ref{fig:econ_text_ori}. The differences between the partisan groups are less pronounced but the results suggest that there is a longer delay for the market to react to a news semantic change from a centrist outlet and that right-wing outlets only have an effect on the market at lags 5-7.

\begin{figure}[t]
    \centering
    \begin{subfigure}[b]{0.75\linewidth}
        \centering
        \includegraphics[width=\linewidth]{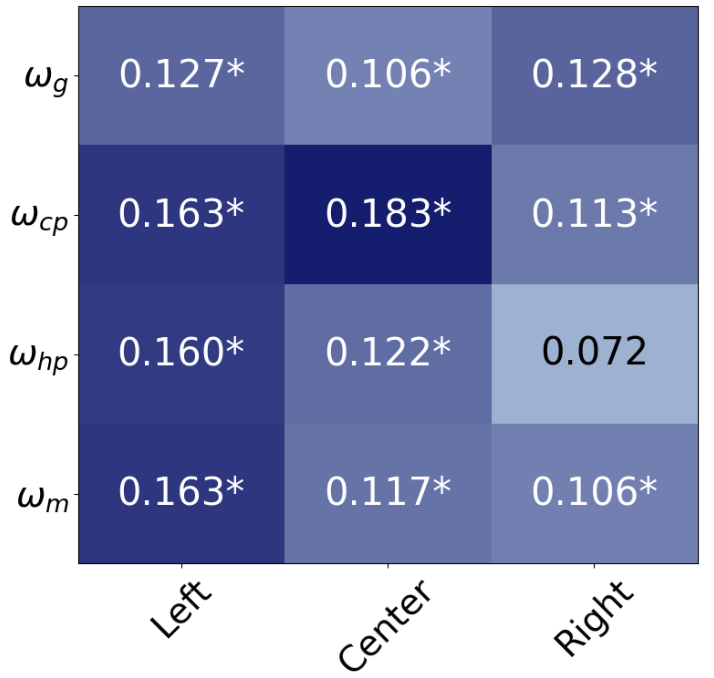}
        \caption{\{$y=\text{text}$, $x=\text{econ}$\}. News outlets responding to market shocks.}
        \label{fig:econ_text_shocks}
    \end{subfigure}
    \vspace{0.5cm} 
    \begin{subfigure}[b]{0.75\linewidth}
        \centering
        \includegraphics[width=\linewidth]{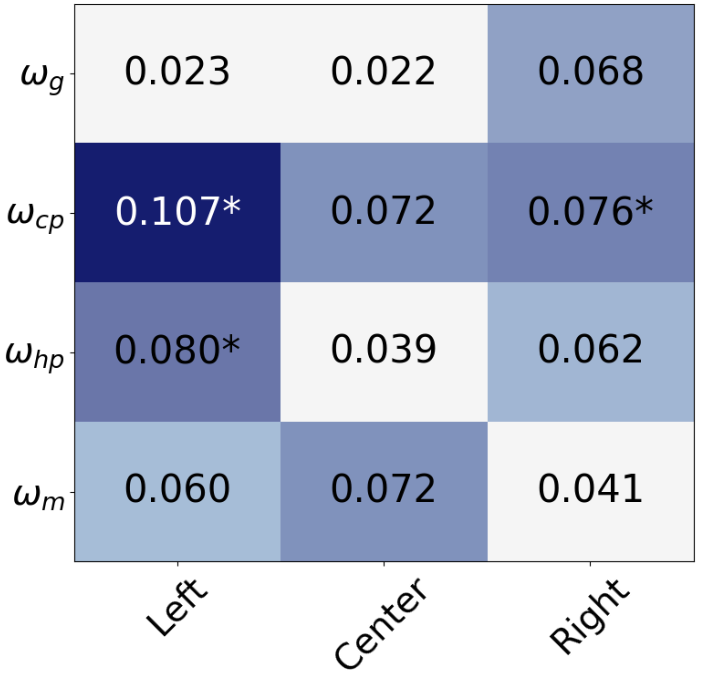}
        \caption{\{$y=\text{econ}$, $x=\text{text}$\}. News outlet changes preceding market shocks.}
        \label{fig:text_econ_shocks}
    \end{subfigure}
    
    \caption{Significant proportion for different market shocks for each political groups across all lags. An * denotes a significant effect inline with \cref{sec:stat_test} and \cref{sec:p_hack}, and the shade of \textcolor{blue}{blue} represents the size of the group effect, with darker colours pertaining to larger effects.}
    \label{fig:combined_shocks}
\end{figure}

Fig. \ref{fig:combined_shocks} shows the proportion of significant pairs for each market shock, aggregated across all lags for each political orientation. It is clear that different political groups react to and precede different market shocks. Similar to Fig. \ref{fig:text_caus_by_ori}, the overall level of significant pairs is high in left-wing news outlets. Fig. \ref{fig:text_econ_shocks} shows that there is not much partisan skew in the proportion of news outlets that respond to growth shocks, but that there are large differences in the response to other market shocks. Centrist outlets are particularly reactive to common risk premium shocks ($\omega_{cp}$) - a shock related to inflation risk. Fig. \ref{fig:econ_text_shocks} shows that left‐wing outlets exhibit a higher proportion of significant pairs for $\omega_{cp}$. The left-wing rate of 10.7\% is significantly higher than centrist and right-wing rates, providing a predictive opportunity for researchers predicting this market shock. It seems that in general news outlets contain more information useful for predicting changes in $\omega_{cp}$ than any other shock. This could be because news about inflation is more likely to influence consumer behaviour because it affects everyone, which in turn affects market movements. In contrast, news outlets of all political groups contained little information useful for predicting growth shocks. 
\vspace{-0.1cm}

\subsection{Confounders}
\label{sec:conf}

Confounders represent variables that cause both the treatment and outcome variables to change \cite{VanderWeele2019}. This can result in false-positive causal pairs being identified. To test whether any of the significant pairs, identified using the Bonferroni correction, were affected by confounding variables, we tested for Granger causality with the other news outlets, market shocks, emotion scores aggregated over all of the news outlets and market volatility. We used the Bonferroni correction to determine the significance level. This resulted in a modest 5.36\% of significant pairs having a confounding variable. The most common confounder type was the news emotion scores; implying that overall emotion can sometimes drive both changes in news discourse and the market. Given the low confounder identification rate, it is clear that for the majority of significant pairs, the mutual interaction is direct. The full analysis is outlined in \cref{app:confounders}.
\vspace{-0.5cm}

\subsection{Feedback Loops}
\label{sec:circular_causality}
Identifying market shock and news outlet pairs that are susceptible to feedback loops can help researchers identify potential viral behaviour.
Circular causality was identified by using the Bonferroni adjusted p-values to infer a significant degree of predictive causality, if both variables in the pair were deemed causal via the methodology in \cref{sec:p_hack} then circular causality was flagged. Although there were not many pairs that exhibited circular causality in our study, the pairs with feedback loops tended to involve $\omega_{cp}$ shocks: 4.22\% of $\omega_{cp}$ pairs displayed feedback loops. Monetary policy $\omega_{m}$ shocks also showed a modest presence of circular dynamics with 1.56\% of all pairs containing feedback loops, exceeding the minimal 0.2\% seen in hedging risk premium $\omega_{hp}$ shocks. Growth-related shocks $\omega_g$ exhibited no feedback loops, which may be explained by the fact that news outlets provide limited information for growth shocks (Fig. \ref{fig:combined_shocks}). 

There are a number of reasons why feedback loops might tend to form with $\omega_{cp}$ shocks. Firstly, inflation-linked shocks often ignite widespread media coverage, amplifying investors’ sensitivity to any hint of price instability. Secondly, frequent and sometimes conflicting potential shock-inducing events linked to inflation -- such as commodity prices, central bank signals, or wage growth -- provide fertile ground for repeated bouts of speculation. Thirdly, repeated episodes of speculation can quickly magnify market moves, drawing additional coverage and further intensifying the original inflation narrative. There are fewer potential shock-inducing events for $\omega_m$ shocks, which could explain the lower amount of feedback loops compared to $\omega_{cp}$ shocks. Hedging premium shocks $\omega_{hp}$ - which arise when markets are concerned with long-horizon cash flow - are more likely to occur in rare episodic crises when there is a demand for safe assets. Fewer shock-inducing events reduce the likelihood of feedback loops. The shock type is also less intuitive, meaning that news may be less likely to cover such shocks.
\vspace{-0.2cm}
\subsection{Temporal Effect}
\label{sec:res_temporal}

\begin{figure}[t]
    \centering
    \begin{subfigure}[b]{1.05\linewidth}
        \centering
        \includegraphics[width=\linewidth]{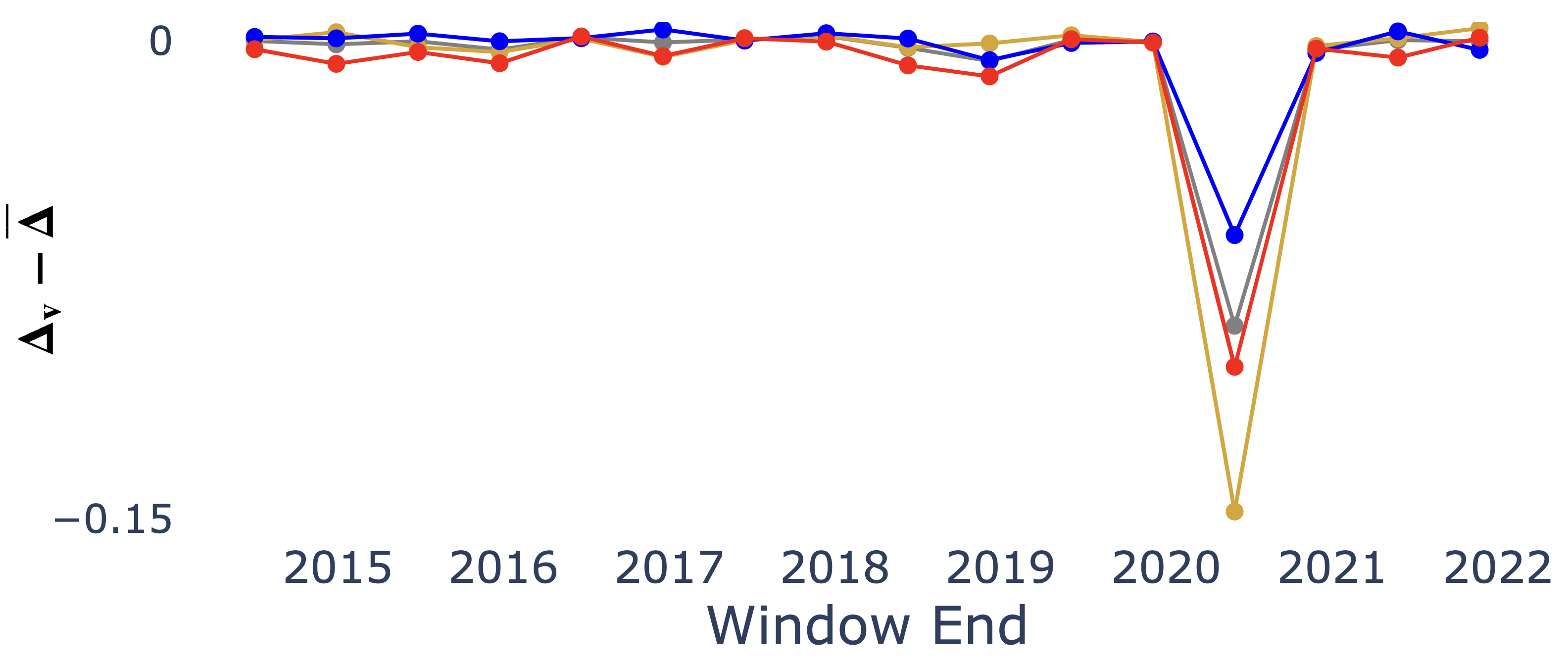}
        \caption{\(\{y=\text{text}, x=\text{econ}\}\)}
        \label{fig:text_econ_temporal}
    \end{subfigure}
    \vspace{0.5cm}
    \begin{subfigure}[b]{1.05\linewidth}
        \centering
        \includegraphics[width=\linewidth]{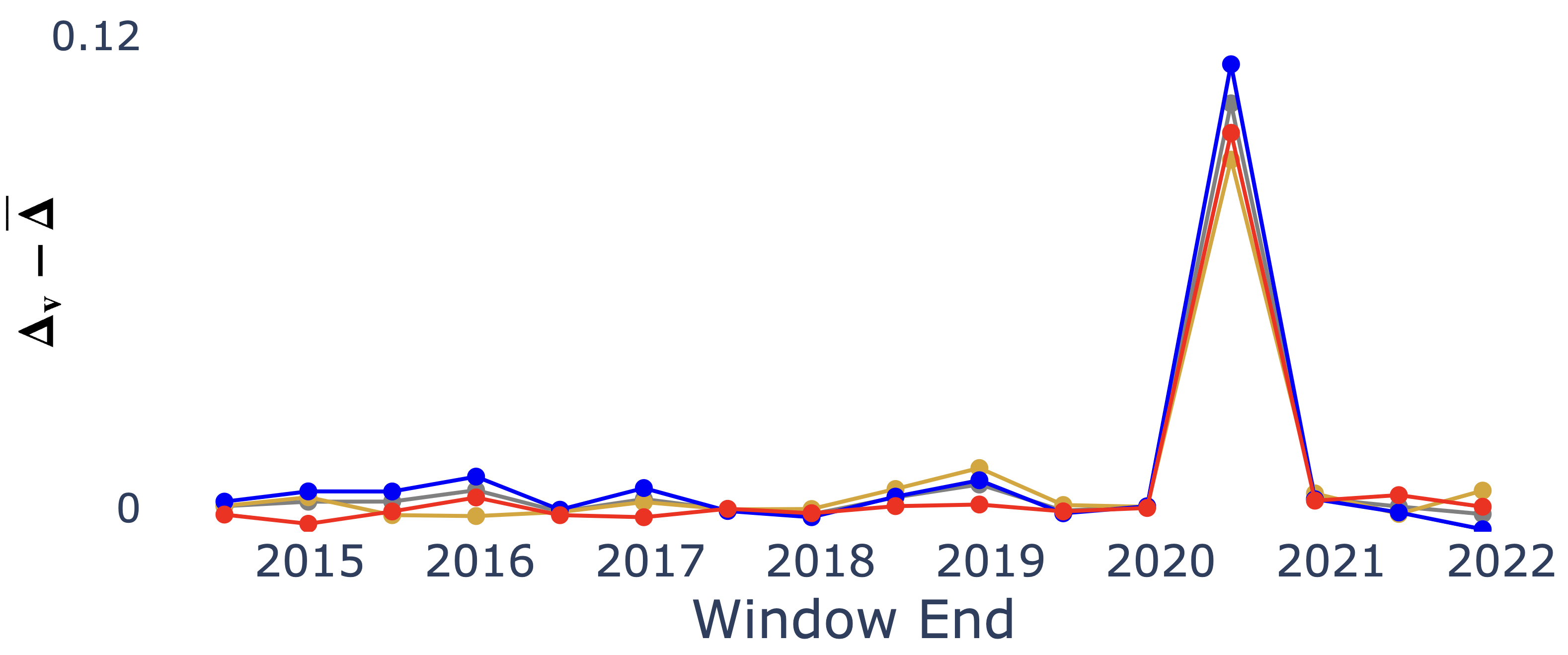}
        \caption{\(\{y=\text{econ}, x=\text{text}\}\)}
        \label{fig:econ_text_temporal}
    \end{subfigure}
    
    \caption{Deviation in MSE improvement of the enhanced model over the base model, \(\Delta_{v} - \overline{\Delta}\), for each test window. Different orientations are indicated with the following colors: \textbf{\textcolor{blue}{left}}, \textbf{\textcolor{red}{right}}, \textbf{\textcolor{Goldenrod}{center}}, and \textbf{\textcolor{gray}{all}}.}
    \label{fig:temporal_MSE}
\end{figure}

To examine how news outlets react to or precede market shocks during different time-periods, we used a rolling evaluation framework. Each window spanned 365 days of data, with the last 90 days in each window forming the test set; consecutive windows were separated by 180 days. Within each window \(v\), let $\text{MSE}_B^v$ and $\text{MSE}_E^v$ be the test-set mean-squared errors of the base (B) and enhanced (E) models, respectively. We define the MSE improvement of model E over B for window \(v\) as \(\Delta_{v}\). Let $\overline{\Delta}$ denote the average improvement across all test windows. Figure~\ref{fig:temporal_MSE} plots the deviation \(\Delta_{v} - \overline{\Delta}\) for each test set. Points above zero represent windows in which the enhanced model’s improvement over the base model exceeded the average improvement, whereas points below zero indicate windows where the enhanced model’s improvement fell short. 

The most striking outlier occurs during the 03/20 to 05/20 test window, highlighting an atypical change in the performance differential between the base and enhanced models. The result shows that information from text improved the ability of models to predict market shocks during COVID-19, whereas market shocks were less useful in predicting changes in news discourse. Financial markets during this time exhibited unusual behaviour, which meant that fewer movements could be endogenously explained using historical market information resulting in more structural shocks. It also suggests that during this time the markets were highly reactive to outside news, showing that text is particularly useful during market events which are being mostly driven by changes exogenous to the market, such as COVID-19. Fig. \ref{fig:text_econ_temporal} also shows that market information was worse at predicting semantic changes in news outlets during early 2020, which could be because news was being driven by the pandemic rather than the market shocks.
\vspace{-0.1cm}
\section{Conclusion}
\label{sec:conclusion}
\vspace{-0.1cm}

In this paper, we find that left-wing and centrist outlets exhibit a stronger and faster response to market changes, whereas right-wing outlets show a smaller effect. Our novel methodology demonstrates, that shifts in semantic embedding space can be causally linked to measurable real-world shocks. By applying a bidirectional Granger-causality framework to text and economic indicators, we uncover how partisanship shapes both the predictability of market shocks and the reactivity of news discourse. We also identify some feedback loops between market shocks and news outlets. Additionally, we provide strong evidence that text-based features serve as essential exogenous variables in unexpected events, such as COVID-19. Our work underscores the value of semantic representations for understanding and anticipating complex economic phenomena.
\vspace{-0.6cm}

\paragraph{Future Work.} Building on our findings, several directions emerge for future investigation. A natural extension involves applying our methodology across international and multilingual contexts, where cultural differences could influence the sensitivity of media discourse to market shocks. Additionally, adopting richer embedding techniques or domain-adapted language models may further illuminate subtle shifts in discourse, particularly during black-swan events such as pandemics or geopolitical crises. Incorporating more granular lag structures that are shorter than 1 day could yield finer temporal insights into the interplay between markets and news discourse. There is also potential to expand our causal framework by integrating unstructured shocks or continuing our confounder analysis in \cref{app:confounders} to create a causal graph that captures interdependencies among multiple shocks and outlets. Finally, evaluating the robustness of our approach in real-time forecasting environments could pave the way for practical applications in finance, policymaking, and risk management.

\section{Limitations}
\label{sec:limitations}

While our methodology establishes a novel framework for linking semantic embedding shifts to real-world financial market shocks, some limitations should be acknowledged. First, although we employed multiple regression models (e.g., linear, kernel-based), further scrutiny with deep learning architectures or ensemble methods could yield different insights. Second, our focus on English-language news articles limits the generalisability to other linguistic or cultural contexts where the relationship between media discourse and markets might follow different dynamics. Third, we focused on identifying the presence and strength of causal relationships and did not cover directionality, representing an opportunity for future research. Finally, we relied on predefined partisan labels and consolidated market shock definitions, but these categorisations can be fluid and may not capture all subtleties in political orientation or economic behaviour.

\section*{Acknowledgments}

The research was funded by the Alan Turing Institute, the UK Defence Science and Technology Laboratory, and the European Research Council (Consolidator Grant Agreement 819131). The first author was funded by the Economic and Social Research Council of the UK via the Grand Union DTP. This work was supported in part by a grant from the Engineering and Physical Sciences Research Council (EP/T023333/1). We are also grateful to the Oxford-Man Institute of Quantitative Finance, the Oxford e-Research Centre, and the Department of Economics for their support.

\bibliography{acl_latex}

\appendix

\section{Linear vs. Non-linear}
\label{app:linvsnonlin}

\begin{figure}[h]
    \centering
    \begin{subfigure}[b]{\linewidth}
        \includegraphics[width=\linewidth]{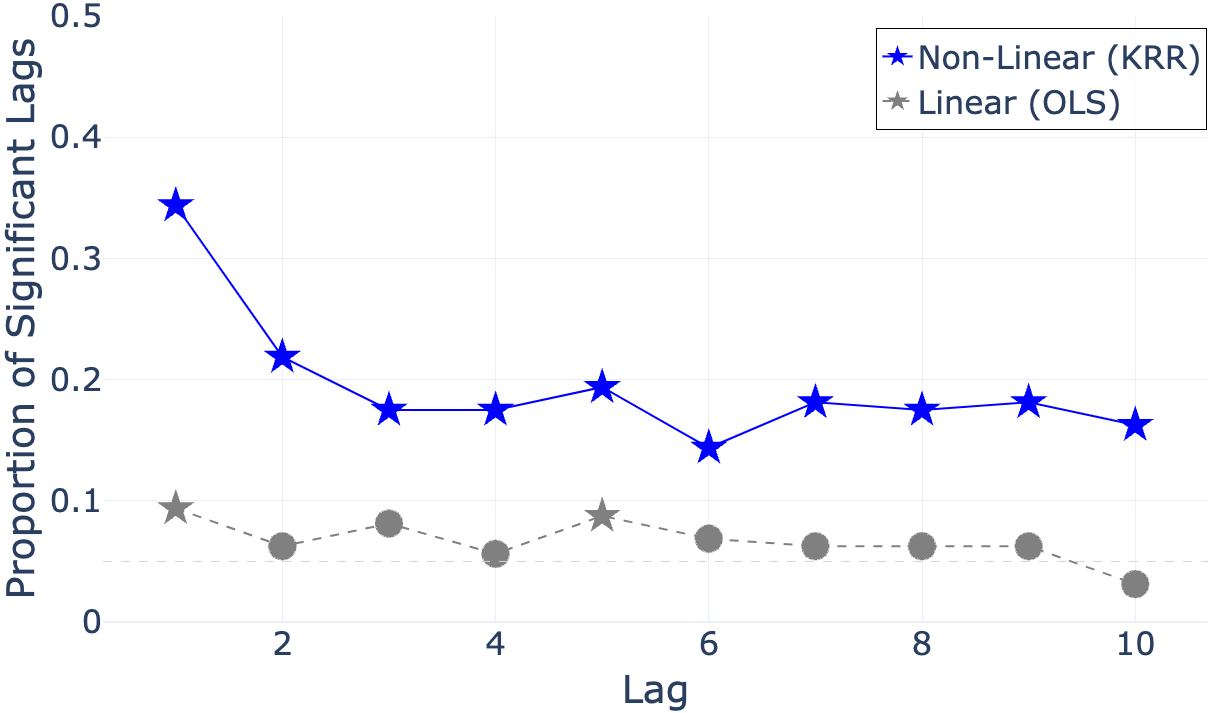}
        \caption{\(\{y=\text{text}, x=\text{econ}\}\)}
        \label{fig:5}
    \end{subfigure}
    
    \vspace{0.5cm}
    
    \begin{subfigure}[b]{\linewidth}
        \includegraphics[width=\linewidth]{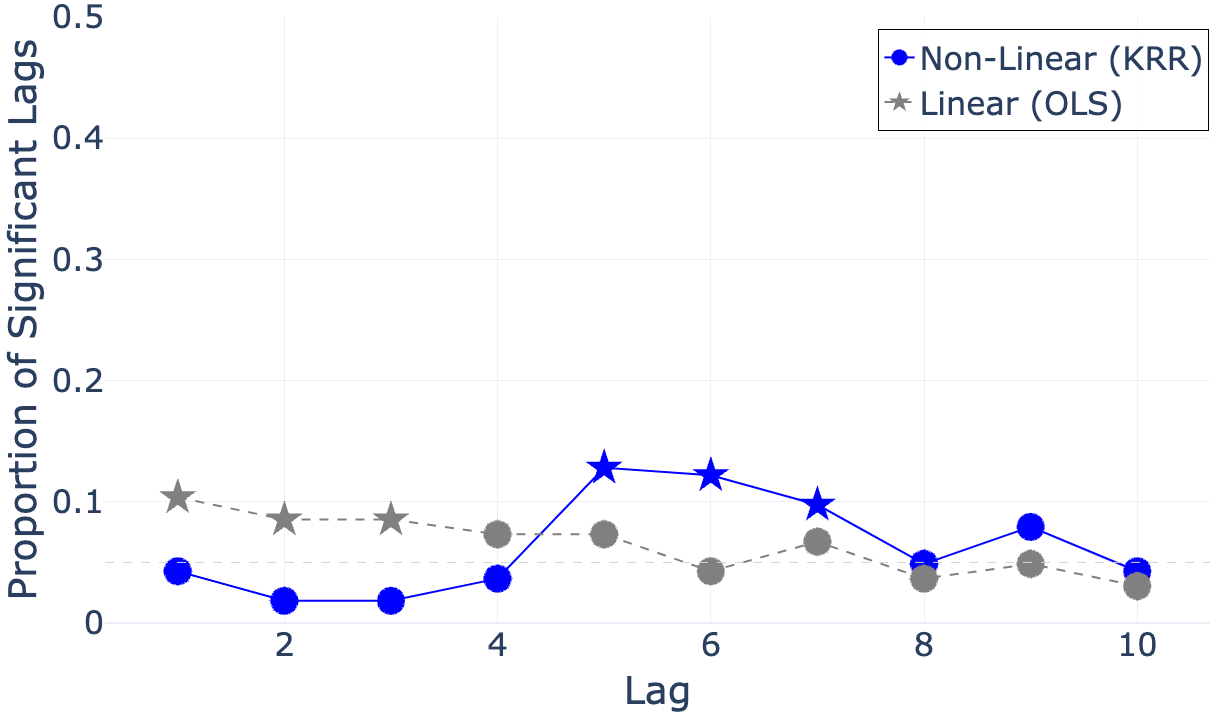}
        \caption{\(\{y=\text{econ}, x=\text{text}\}\)}
        \label{fig:6}
    \end{subfigure}
    
    \caption{The proportion of significant pairs when \{$y=\text{text}$, $x=\text{econ}$\} and \{$y=\text{econ}$, $x=\text{text}$\} for all news outlets. Comparing the \textbf{\textcolor{blue}{Non-linear}} and \textbf{\textcolor{gray}{Linear}}.}
    \label{fig:combined3}
\end{figure}

Tab. \ref{fig:combined3} displays a more granular comparison between the linear and non-linear models outlined in \cref{sec:reg_models}. 

\section{News Outlets}
\label{app:news_outlets}

We use the articles from the following outlets:

\begin{itemize}
    \item \textbf{Center:}  
    \begin{itemize}
        \item Heavy, CNBC, Business Insider, Reuters, Wall Street Journal, USA TODAY, BBC News, Forbes, FiveThirtyEight
    \end{itemize}
    
    \item \textbf{Left:}  
    \begin{itemize}
        \item The New Yorker, Newsweek, CNN, The New York Times, CBS News, Vox, Rolling Stone, The Washington Post, The Guardian, Slate, Daily Beast, Vice, Politico, HuffPost, MSNBC, ABC News
    \end{itemize}
    
    \item \textbf{Right:}  
    \begin{itemize}
        \item The Gateway Pundit, PJ Media, The Western Journal, Breitbart News, Washington Times, ZeroHedge, New York Post, Reason, Fox News, Newsmax, Red State, The Epoch Times, Townhall, Orange County Register, The Daily Caller, National Review, Bizpac Review, TheBlaze
    \end{itemize}
\end{itemize}

The political categories come from \citet{spliethoever2022}, which used \href{https://www.allsides.com}{AllSides.com} tags.

\section{SVAR Framework}
\label{app:SVAR}

To generate the financial market shock time series, \cite{cieslak2021common} fit an unrestricted vector autoregression on the daily changes in bond yields (2-year, 5-year, and 10-year maturities) and daily log stock returns. Let \(\mathbf{y}_t \in \mathbb{R}^{4}\) denote the vector of observed variables on day \(t\):
\[
\mathbf{y}_t = 
\begin{pmatrix}
\Delta y_{2\text{yr},t}\\[6pt]
\Delta y_{5\text{yr},t}\\[6pt]
\Delta y_{10\text{yr},t}\\[6pt]
\Delta \ln(S_t)
\end{pmatrix}
\]
where \(\Delta y_{m\text{yr},t}\) is the daily yield change at maturity \(m \in \{2,5,10\}\) and \(\Delta \ln(S_t)\) is the daily log return of the stock index \(S_t\). The structural VAR of lag order \(L\) can be written in the following form:
\[
\mathbf{y}_t = \boldsymbol{\mu} 
+ \sum_{j=1}^{L} A_j \,\mathbf{y}_{t-j} 
+ \mathbf{u}_t
\]
where \(\boldsymbol{\mu} \in \mathbb{R}^4\) is a constant term, each \(A_j \in \mathbb{R}^{4\times 4}\) is a coefficient matrix for lag \(j\), and \(\mathbf{u}_t \in \mathbb{R}^4\) is the reduced-form error term with covariance matrix \(\Sigma = \mathbb{E}[\mathbf{u}_t \mathbf{u}_t^\top]\).

\vspace{1em}
\paragraph{Identifying Structural Shocks.}
Following standard practice for structural identification, the covariance matrix \(\Sigma\) is factorised, with a lower-triangular Cholesky decomposition:
\[
\Sigma = P P^\top
\]
where \(P \in \mathbb{R}^{4\times 4}\) is invertible and maps the structural shocks \(\boldsymbol{\varepsilon}_t \in \mathbb{R}^4\) to the reduced-form errors:
\[
\mathbf{u}_t = P \,\boldsymbol{\varepsilon}_t,
\quad
\boldsymbol{\varepsilon}_t \sim \mathcal{N}(\mathbf{0}, I_4)
\]
Sign restrictions \cite{fry2011sign} are then imposed on the impulse responses, which amounts to applying an orthonormal rotation \(Q\) to \(P\), yielding
\[
\mathbf{u}_t = P\,Q\,\widetilde{\boldsymbol{\varepsilon}}_t
\]
where \(Q \in \mathbb{R}^{4\times 4}\) satisfies \(QQ^\top = I_4\) and \(\widetilde{\boldsymbol{\varepsilon}}_t\) are the rotated structural shocks. Among all permissible rotations, the impulse whose response satisfied the sign restrictions for each of the four shocks (\emph{growth}, \emph{monetary}, \emph{common premium}, and \emph{hedging premium}) is selected. 

\section{Confounders Analysis}
\label{app:confounders}

\begin{figure*}[ht]
    \centering
    \includegraphics[width=\linewidth]{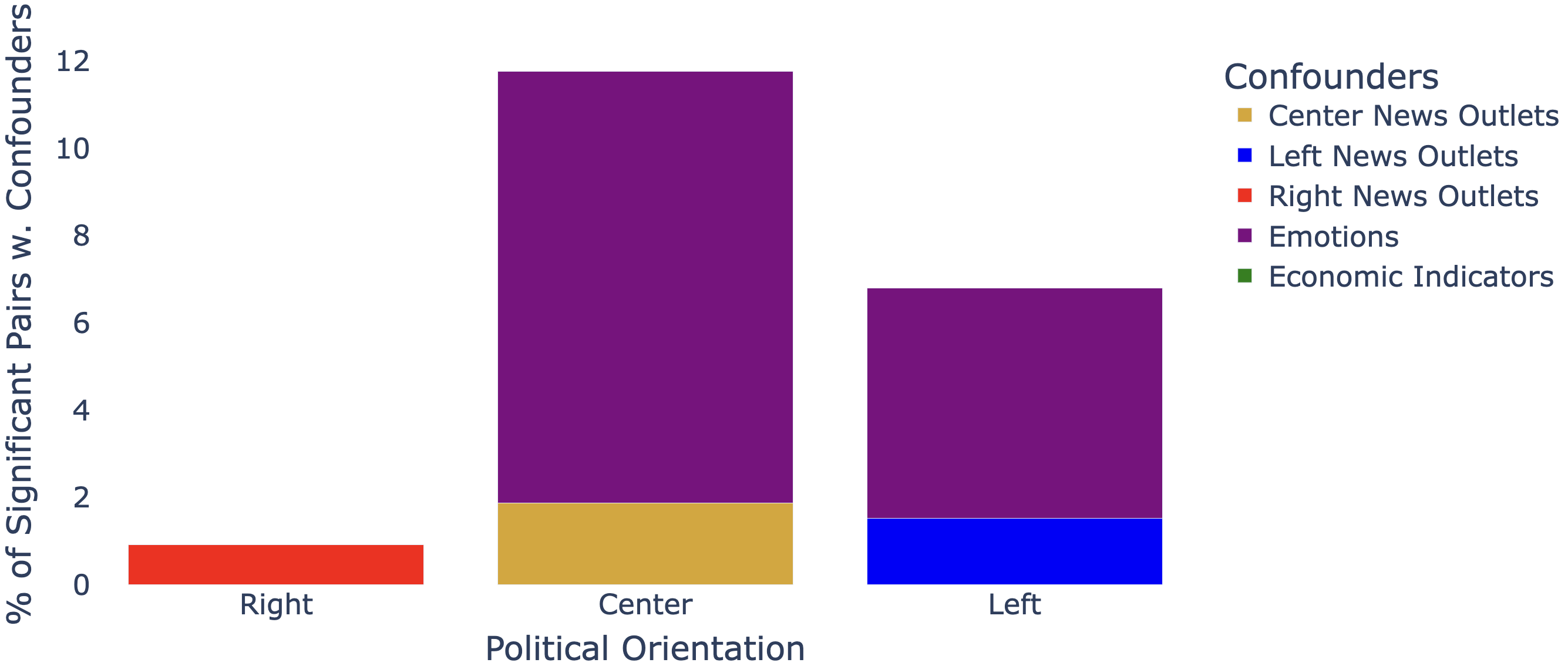}
    \caption{Proportion of significant pairs where \{$y=\text{text}$, $x=\text{econ}$\} that have a confounding variable by orientation. The category of the confounding variables are denoted by the colour of the bar section.}
    \label{fig:confounders}
\end{figure*}

To understand the nature of the significant pairs that were identified using the Bonferroni correction, we tested whether the causal pairs had any confounding variables that caused both the treatment and outcome variables. The confounders were identified by whether a single variable significantly improved the performance of both the treatment ($x$) and outcome ($y$) variables using a Bonferroni correction. The following variables were tested: the cosine distance of all of the other news outlets, the other structural shocks, VIX volatility index (which is grouped with the Economic Indicators in Fig. \ref{fig:confounders}), and an emotion score for the news in general. The emotion score was calculated by first applying a \href{https://huggingface.co/j-hartmann/emotion-english-distilroberta-base}{pre-trained emotion classification model} to each news article, which assigned probability scores to a set of predefined emotion labels. For each day \( d \), the emotion score for label \( \ell \) was computed as the average predicted probability across all articles published on that day:

\[
E_{\ell, d} = \frac{1}{N_d} \sum_{i=1}^{N_d} p_{i}(\ell),
\]

where \( p_i(\ell) \) is the predicted probability of emotion \( \ell \) for article \( i \), and \( N_d \) is the total number of articles published on day \( d \). This aggregation produced a time series of daily average emotion scores, which were then tested for their role as potential confounders in the causal relationships between the treatment and outcome variables. The potential confounding variables had to have a reporting frequency of 1 day which limited the total number of variables that were trialled.

Fig. \ref{fig:confounders} shows the resulting percentage of total significant pairs that had a confounder. Firstly, there are no examples where there is a confounding news outlet from another partisan group, which shows that each partisan group has limited influence on outlets with different political outlooks. Secondly, there were no economic indicators that were confounding variables largely due to the fact that none of the economic indicators were deemed as causal for the other economic indicators. This could be explained by the Efficient Market Hypothesis, which states asset prices can be explained by all available information, meaning that the other economic variables do not add any predictive information. What is interesting is that the news emotion score is often a confounding variable, causing changes in both economic indicators and the news outlets. This suggests that not everything is priced in and that market shocks and news outlets can react to overall emotions within the news. However, right-wing outlets do not react to news emotion scores, showing a distinct difference between partisan groups.
\vspace{-0.5cm}
\section{Licenses}
\label{app:licenses}
\vspace{-0.3cm}

The news data from \citet{spliethoever2022} was not published with a license, but the original source of the data - \href{https://commoncrawl.org/news-crawl}{Commoncrawl News} - has a permissive Apache 2.0 license. The market shock data from \citet{cieslak2021common} was shared by the authors. 
\vspace{-0.5cm}

\section{Model Setup}
\label{app:models}
\vspace{-0.1cm}

To simplify the experimental setup and limit the number of observations that could distort the statistical significance of our findings we used the default parameters for both the Linear and KRR models, except we used an \textit{RBF} kernel instead of a \textit{linear} kernel for the KRR regressor. If we had done hyperparameter tuning, then the statistical significance corrections would have been more strict. We used the \href{https://scikit-learn.org/stable/index.html}{sklearn} package for both regressors, using the \href{https://scikit-learn.org/stable/modules/generated/sklearn.linear_model.LinearRegression.html}{LinearRegression} and \href{https://scikit-learn.org/stable/modules/generated/sklearn.kernel_ridge.KernelRidge.html}{KernelRidge} modules.
\vspace{-0.1cm}

\section{Dataset Statistics}
\label{app:data_stats}
\vspace{-0.1cm}

\begin{figure}[b!]
    \centering
    \includegraphics[width=\linewidth]{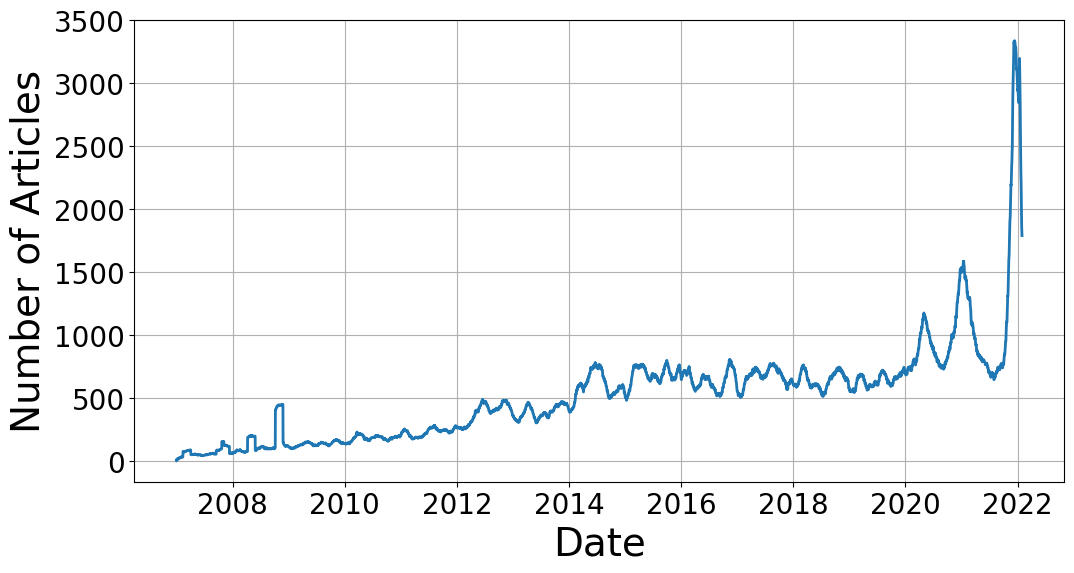}
    \caption{30-day rolling average of articles by day.}
    \label{fig:daily_counts}
\end{figure}

Fig. \ref{fig:daily_counts} shows the 30-day rolling average of article counts after the filtering outlined in \cref{sec:datasets}. The spike in 2022 can be attributed to the fact that \citet{spliethoever2022} used a web scrape from 2022 to create the dataset. The data came from a \href{https://commoncrawl.org/news-crawl}{Commoncrawl News} WARC file, which contains the news available on the day of the web scrape. While the publish date is being displayed on Fig. \ref{fig:daily_counts}, older articles might not be available. The absolute number of articles is not overly important so long as there is sufficient coverage. Cosine distance means that there is a value for every day - including a value of zero on days where no articles were published.

The outlet counts are displayed in Tab. \ref{tab:article_counts}. This represents all of the articles from 01/07 to 02/22. Some of the outlets should evidently have more articles, however companies like \href{https://www.nytimes.com/2023/12/27/business/media/new-york-times-open-ai-microsoft-lawsuit.html}{The New York Times} have recently attempted to combat web scraping by restricting access to bots. To make sure that we have sufficient information contained in the text-based time-series, we only consider outlets which have published articles on 25\% of the days, including the rolling window framework.

\begin{table}[b]
    \centering
    \begin{tabular}{|l|r|}
        \hline
        \textbf{Outlet} & \textbf{Article Count} \\
        \hline
        Vox & 382,666 \\
        Newsweek & 246,740 \\
        HuffPost & 192,210 \\
        CNBC & 185,890 \\
        Daily Beast & 131,291 \\
        Reuters & 130,678 \\
        Breitbart News & 129,580 \\
        Slate & 125,184 \\
        Wall Street Journal & 123,964 \\
        Washington Times & 113,836 \\
        Newsmax & 108,015 \\
        New York Post & 106,676 \\
        Vice & 92,927 \\
        The Washington Post & 90,056 \\
        The New Yorker & 89,003 \\
        CNN & 76,320 \\
        PJ Media & 65,282 \\
        Rolling Stone & 52,896 \\
        The Guardian & 44,671 \\
        USA TODAY & 38,355 \\
        ABC News & 37,414 \\
        MSNBC & 35,942 \\
        The Epoch Times & 30,693 \\
        Red State & 28,435 \\
        Townhall & 26,345 \\
        Forbes & 25,488 \\
        CBS News & 23,202 \\
        Orange County Register & 22,529 \\
        BBC News & 21,683 \\
        National Review & 20,297 \\
        ZeroHedge & 18,497 \\
        Heavy & 16,618 \\
        Bizpac Review & 16,492 \\
        Reason & 16,407 \\
        FiveThirtyEight & 16,036 \\
        The Daily Caller & 15,768 \\
        TheBlaze & 10,778 \\
        Fox News & 10,133 \\
        Business Insider & 9,911 \\
        The Western Journal & 9,666 \\
        The Gateway Pundit & 7,240 \\
        Politico & 5,468 \\
        The New York Times & 5,021 \\
        \hline
    \end{tabular}
    \caption{Article counts per news outlet}
    \label{tab:article_counts}
\end{table}

\end{document}